\documentclass[12pt]{article}
\usepackage[margin=1in]{geometry}
\usepackage[T1]{fontenc}          
\usepackage[utf8]{inputenc}       
\usepackage{graphicx}
\usepackage{amsmath,amssymb,amsfonts}
\usepackage{booktabs}
\usepackage{enumitem}
\usepackage{caption}
\usepackage[nointegrals]{wasysym} 
\usepackage{placeins}
\usepackage{microtype}
 \usepackage{stix2}
\usepackage[authoryear,round]{natbib}   
\usepackage[title]{appendix}
\usepackage{authblk}                    
\usepackage{hyperref}                   

\begin{document}
 
\title{The Theory of Mind Utility\\[0.35em]
       \large A Formal Account of Mentalizing}
 
\author[1]{Nikolos Gurney\thanks{Corresponding author:
  \href{mailto:gurney@ict.usc.edu}{gurney@ict.usc.edu}}}
\author[2]{Stacy Marsella}
 
\affil[1]{Institute for Creative Technologies, University of Southern
  California, 12015 E Waterfront Dr, Playa Vista, CA 90094, USA.\\
  ORCID: \href{https://orcid.org/0000-0003-3479-2037}{0000-0003-3479-2037}}
\affil[2]{Khoury College of Computer Sciences, Northeastern University,
  440 Huntington Avenue, Boston, MA 02115, USA.\\
  ORCID: \href{https://orcid.org/0000-0002-5711-7934}{0000-0002-5711-7934}}
 
\date{}
 
\maketitle
 
\begin{abstract}
Inferring another person's beliefs requires reconstructing their
information access history: what they encountered, in what order, from
whom, and with what credibility. Existing formal accounts of theory
of mind generally treat beliefs as given. We introduce the Theory of
Mind Utility (ToM-U), a computational-level theory of how beliefs are
formed from information-access history. ToM-U represents a person's
epistemic situation as a Local Epistemic World Model (LEWM): a
directed typed graph linking agents, information, and belief-like
states. ToM-U generates candidate LEWMs, limits how deeply a person
reasons about another's reasoning, and evaluates each candidate
against observed behavior. A sufficiently coherent candidate is
accepted, while a failed one leaves a structured residue that shapes
later reasoning. Together, these components formalize a distinct
cognitive process that operates across social reasoning. The
architecture supports testable predictions about when mentalizing will
fail and how those failures will affect subsequent attempts.
\end{abstract}
 
 
\maketitle

\section{Introduction}
\label{sec:intro}

Inferring the mental states of others, or mentalizing, is fundamental 
to social life. Contemporary psychology recognizes this ability as 
having a \textit{theory of mind} (ToM), a construct that has its modern 
foundations in Premack and Woodruff's seminal work on chimpanzees' 
ability to reason about each other's mental states 
\citep{premack1978does}. The mental states that ToM infers and reasons 
over extend beyond knowledge or informational representations; they 
include stances, like beliefs, held in the mind concerning the targets 
of cognition. For example, the canonical empirical test of ToM is 
focused on reasoning about \emph{beliefs} and why people hold them 
\citep{wimmer83}. The crux of inferring beliefs, or any similar mental 
state, is an epistemic provenance problem: if Sam sees a bag labeled 
``chocolate'' and has no further information, she will conclude it 
contains chocolate; if before she saw the label a friend said, ``that 
bag contains popcorn,'' Sam would likely believe it contains popcorn 
\citep{ullman23}. In either case, having ToM for Sam's belief about the 
bag requires answering questions related to what Sam was exposed to, in 
what order, from whom, and how credibly. The ToM problem is one of 
epistemic state inference based on \textit{information access history}. 
The \textit{Theory of Mind Utility} (ToM-U) is a social reasoning theory 
presented as a formal framework describing how minds solve this problem. 

The ToM-U approach differs significantly from past ToM formalizations. 
The successful Bayesian Theory of Mind (BToM) approach formalizes goal 
and belief inference as rational inverse planning, typically over 
Markov decision processes \citep{baker09,baker2011bayesian}. Unlike 
ToM-U, BToM generally takes belief states as inputs rather than deriving 
them from information-access history and source credibility. Moreover, 
BToM handles false beliefs by manipulating perceptual access, whereas 
the ToM-U approach addresses the more general problem of belief 
formation based on information exposure and its modifiers like trust. 
BToM also relies on planning models as the representational object, 
whereas ToM-U implements an epistemic graph. Similarly, adjacent 
computer-science research uses action and utility as its 
representational foundation, leaving belief formation from epistemic 
states outside its 
scope~\citep{pynadath2005psychsim,pynadath2007minimal}. 

ToM-U also differs in important ways from existing representational  
accounts of how humans accomplish mentalizing. In simulationism, the 
self is the generative engine: we understand others' mental states by 
simulating them in our own minds, rather than using a theoretical, 
rule-based approach \citep{gordon1986folk,goldman2006simulating}. This 
yields candidate mental states constrained by the perspective of the 
person doing the reasoning. In contrast, ToM-U does not take the 
mentalizer's own perspective as the model. ToM-U constructs candidate 
mental states from the outside, constrained by the (known or assumed) 
information access history of the mentalizing target, so that the 
candidates reflect what is available to the target from the target's 
epistemic position, not what the mentalizer would believe in the 
target's place. 

A leading alternative to simulationism, theory-theory, argues that 
people understand minds by forming ``naïve'' or ``folk'' theories, much 
like scientists form hypotheses \citep{gopnik92,wellman1990child}. 
Theory-theory posits tacit rules about agent behavior but leaves the 
representational substrate for epistemic inference unspecified. ToM-U 
provides a formal mechanism that takes ordered information-access 
history, source credibility, and observability constraints as inputs 
and produces a belief-state estimate.

ToM-U posits the generation of \textit{Local Epistemic World Models} 
(LEWMs) to achieve mentalizing. Conceptually, LEWMs are similar to 
mental models in that they are representations constructed on the fly to 
support specific inferential tasks \citep{johnson83}. They differ, 
however, in their formalism. A LEWM is a directed typed graph 
constructed on the fly, evaluated for coherence against observed 
behavior, and accepted or rejected as a working account of another 
agent's epistemic situation. This construction, evaluation, and judgment 
process combined with a means of handling lingering effects of past 
LEWMs constitutes ToM-U. 

The foundational positions of ToM-U that give rise to its 
formal structure include: 
\begin{itemize}
    \item Belief states are inferred from ordered exposure history 
    and source credibility. The sequence and provenance of 
    information access, not only its content, determine the epistemic 
    state attributed to the target. 
    \item Inference proceeds by generating and evaluating discrete 
    candidate world models.
    \item Recursive mentalizing is represented as a bounded branching 
    tree. The focal agent's sophistication limits the overall recursive 
    depth and branching width it can construct, while its estimate of 
    each target agent's sophistication limits the recursive depth 
    attributed to that target. At each depth, the represented agent is 
    treated as a distinct constructed entity rather than as the level-0 
    version of that agent with adjusted inputs. This architecture 
    prevents infinite regress without requiring additional theoretical 
    commitments.
    \item Failed mentalizing attempts leave a structured trace that 
    impacts future inference (a ``residue''). 
\end{itemize}
These architectural commitments bound ToM-U while keeping its 
representational and inferential machinery independent of any 
particular social domain. Moreover, because epistemic states are the 
theorized output of ToM-U, these commitments set it up to serve as an 
input to downstream processes. For example, belief states output from it 
could feed into a BToM-style goal inference process. 

Important contributions of the Theory of Mind Utility are: 
\begin{itemize}
    \item Formally specifying the epistemic state inference problem that 
    is inherent to mentalizing such that it lends itself to empirical 
    validation. 
    \item Introducing local epistemic world models as typed graphs that 
    capture ordered information access history as their representational 
    substrate. 
    \item A generate-and-filter inference architecture with bounded 
    sophistication and explicit confidence accumulation. 
    \item A mechanism (the residue function) by which failed mentalizing 
    attempts leave a persistent trace that shapes subsequent reasoning 
    events. 
\end{itemize}
ToM-U is a computational-level theory: it specifies the function, inputs, 
outputs, and rationale of epistemic-state inference without selecting an 
algorithmic or neural implementation \citep{marr82}. The accompanying 
executable model serves as a consistency witness demonstrating that the 
specification is coherent and instantiable.

Together, these contributions specify epistemic-state inference as a 
distinct cognitive problem that existing accounts presuppose but do 
not model. The architecture also identifies conditions under which 
mentalizing should fail and how failed attempts should influence later 
inference.

\section{The Theory of Mind Utility}
\label{sec:overview}

The Theory of Mind Utility describes a domain-agnostic mentalizing 
mechanism. It is engaged whenever a parent process invokes theory of 
mind. Its sole function is constructing candidate models, which we call 
local epistemic world models, in service of belief-like state inference. 
Other functions, such as action prediction, planning, and behavioral 
responses, are explicitly placed as downstream consumers of ToM-U 
outputs. The core functionality of the mechanism relies on generation of 
truncated models and assessing their viability as accounts of reality. 
Accordingly, ToM-U is presented as five definitions covering world model 
representation, agent node structure, proliferation mechanism, inference 
procedures, and residue function. 

\subsection{Illustrative Example: The Popcorn Bag}
\label{sec:popcorn}

A salient example of the unexpected contents task~\citep{perner1987three} 
from recent work investigating the ToM reasoning abilities of 
large language models~\citep{ullman23,kosinski24} serves to illustrate 
ToM-U in operation. The scenario is more transparent as an illustration 
than traditional false belief tasks, where the physical staging of the 
scenario can introduce confounds irrelevant to the epistemic inference 
problem ToM-U addresses.

\begin{quotation}
A bag contains popcorn. The label on the bag reads ``chocolate.'' Sam has
never seen the bag before and cannot see inside it. Her only available
information is the label. A straightforward theory of mind question is:
what does Sam believe is in the bag? The correct answer, grounded in 
Sam's information access history, is chocolate: the label is her only 
information source and she has no reason to distrust it.
\end{quotation}

A slightly more demanding variant introduces a trusted friend who told 
Sam beforehand that the bag contains popcorn and that the label should be
ignored. Sam believes her friend, but she reads the label anyway. The
accurate theory of mind answer is now popcorn; the trusted testimony
overrides the label because it carries higher credibility. Systems that
fail this variant, as some language models have been shown to 
do~\citep{ullman23}, fixate on the most recent or salient surface signal 
rather than maintaining a coherent belief state for Sam that properly 
tracks her full information access history. We argue that the ToM-U 
approach handles this naturally.

\subsection{Motivation: Local Epistemic World Models 
as Uncertainty Resolution}
\label{sec:motivation}

Theory of mind engagement is fundamentally a response to uncertainty
about the mental states of others. Such uncertainty results from 
situational or analytical information being insufficient to explain 
observed behavior. Resolving the uncertainty forces an inference task: 
what belief states, intentions, or attitudes must the other agent hold 
for their behavior to be coherent? We refer to the agent whose 
mentalizing is being modeled as the \textit{focal agent} and the subject 
of the focal agent's mentalizing as the \textit{target agent}. A  
\textit{local epistemic world model} (LEWM) is a representation 
(specifically, a directed typed graph) that is occasioned by a 
particular inferential need and anchored to the focal agent's 
perspective. Its content is restricted to agents, the state nodes 
representing objects, situations, and conditions relevant to the 
occasion, and the epistemic relationships among them, including beliefs, 
informational channels, and credibility judgments. The LEWM is the 
representational structure through which the inference problem is posed 
\textit{and} resolved.

LEWMs draw on both the mental-model tradition \citep{johnson83} and 
situation modeling \citep{kintsch1978toward,vandijk1983strategies}. 
Like those representations, a LEWM is occasion-specific, constructed for 
a particular inferential task, and evaluated for coherence. Its
distinctive contribution is to formalize that representation as a
directed typed graph, making its components and inferential role
available for explicit specification and testing.

Here, ``world model'' refers to a structured hypothesis constructed,
evaluated, and updated for a particular inferential task, rather than
to the reinforcement-learning sense of a learned, persistent model of
environment dynamics \citep{ha18}. ``Epistemic'' restricts its content
to knowledge- and belief-like relationships and their acquisition,
credibility, and justification. ``Local'' means that the model is
occasion-specific, anchored to the focal agent's perspective, and
bounded to the agents and relationships relevant to the current
inference.

A LEWM is a focal agent’s occasion-specific hypothesis about the 
epistemic state of the world, evaluated against observed behavior and 
accepted or rejected as a viable account of reality. This distinguishes 
the LEWM approach from purely reactive signal processing: in lieu of 
responding to surface features of observed behavior, the focal agent 
constructs a candidate representation and asks whether that 
representation makes the behavior intelligible.

Central to the LEWM are the epistemic relationships encoded in its 
edges, the belief-like states (BLSs). The qualifier ``like'' signals 
agnosticism about whether these attributed states meet the full 
conditions for propositional belief in the philosophical sense. BLS  
encompasses attitudes such as credences, dispositions, and 
representational states that fulfill a belief-like functional role 
without requiring stronger conditions on intentionality or content. 
This agnosticism echoes Dennett's (\citeyear{dennett1987intentional}) 
intentional stance, in which belief attribution is a practical 
interpretive act. Full metaphysical commitment would impose stricter 
requirements on the edges and fundamentally change ToM-U. 

Accepting a LEWM therefore implies accepting its BLSs as valid. 
Each BLS is a load-bearing element of a mentalizing session. A LEWM 
in which Sam believes the bag contains chocolate is a fundamentally 
different hypothesis about the state of the world than one in which 
Sam believes it contains popcorn, even if all other elements are 
identical. The inference procedures of ToM-U (Definition 
4~\S\ref{def:inference}) therefore evaluate LEWMs as candidate 
solutions to an uncertainty-resolution problem, accepting whichever 
model best explains observed behavior given available information 
access histories and observability constraints.

Because the focal agent selects among discrete candidate LEWMs rather
than optimizing over a continuous distribution of BLS values, ToM-U
adopts a generate-and-filter inference architecture. Rejection is
itself treated as informative about the epistemic edges implicated in
the failed candidate; Definition~5 (\S\ref{def:residue}) formalizes how
that information shapes later inference.

\section{Formalization of the Theory of Mind Utility}
\label{sec:formlization}

ToM-U is specified through five definitions. Together, they fix the 
representational and inferential constraints that an implementation must 
satisfy while leaving implementation-specific functional forms open.

\textbf{Definition 1} specifies the LEWM, the directed typed graph that
serves as the representational substrate for all subsequent operations.
\textbf{Definition 2} specifies the agent node structure, defining the
internal properties of each agent represented in the LEWM, including
information access history, sophistication, and temporal anchoring.
\textbf{Definition 3} specifies the proliferation mechanism by which 
the focal agent recursively instantiates projected agent nodes, 
constructing a bounded branching tree that represents nested 
mentalizing. \textbf{Definition 4} specifies the three inference 
procedures (backward inference, self-projection, and mutual 
reconciliation) that operate over the proliferated LEWM to produce 
belief-like state estimates. \textbf{Definition 5} specifies the 
residue function, which captures the informational trace left by 
rejected LEWMs and feeds into downstream reasoning processes. 
Together, these definitions constitute the Theory of Mind Utility.

\paragraph{A Note on Notation}
\label{sec:notation}

The formalization combines conventions from graph theory, set theory,
theoretical computer science, and cognitive science. Angle brackets
\(\langle\cdot\rangle\) denote tuples; uppercase Roman letters denote
sets and lowercase subscripted forms their elements; and
\(f:X\to Y\) denotes a function from domain \(X\) to codomain \(Y\).
The calligraphic \(\mathcal{T}\) denotes a vocabulary or type set, while
\(\mathbb{N}\) and \(\mathbb{R}\) denote the natural and real numbers.

For example,
\(W=\langle A,O,E,\text{BLS},\text{obs},\text{cred}\rangle\)
identifies a LEWM by six components. BLS-typed edges are directed,
typed, and weighted: the type label identifies the attitude represented
by the edge, while the scalar supplies its weight.

\paragraph{Preliminary Definitions}
\label{sec:prelim}

The following quantities appear across multiple definitions and are
collected here to avoid forward dependencies. Quantities that appear only
within a single definition are introduced inline. 

\textbf{Observed behavior} \(\beta_i\) denotes the observable actions,
signals, or outputs of target agent \(a_i \in A \setminus \{a_1\}\) that
serve as the primary input to inference. In the multi-agent case, each
target agent has their own observed behavior \(\beta_i\), and inference
procedures are indexed accordingly. In the single-agent case this reduces
to a scalar \(\beta\) for the sole target agent. \(\beta_i\) is treated 
as given; the processes generating it are external to ToM-U.

\textbf{Fit score} \(f_j \in [0,1]\) is a scalar measure of how well the
\(j\)-th candidate BLS configuration accounts for observed 
behavior \(\beta_i\) of target agent \(a_i\), given that agent's 
information access history \(H_i\) and observability constraints 
on relevant edges. The candidate index \(j\) is distinct 
from the target agent index \(i\) used elsewhere. \(f_j = 1\) indicates 
a perfect account; \(f_j = 0\) indicates no explanatory purchase. 

\textbf{Confidence} \(C_\ell \in [0,1]\) is a cumulative weight 
reflecting how well the generate-and-filter procedure has accounted for 
observed behavior after \(\ell\) iterations. Its properties are: 
\(C_0 = 0\); \(C_\ell\) is monotonically non-decreasing in \(\ell\) and 
bounded above by 1; and \(C_\ell \geq \max_{j \leq \ell} f_j\), so that 
confidence is at least as high as the fit of the best candidate evaluated 
so far. One accumulator satisfying these properties, used in the reference 
implementation, is:
\begin{equation}
C_\ell = 1 - \prod_{j=1}^{\ell}(1 - f_j)
\end{equation}
Under these properties, a single high fit score produces rapid 
confidence accumulation, while consistently low fit scores produce slow 
accumulation bounded by the sophistication ceiling.
The scalar accumulator aligns with empirical evidence that human 
judgment relies on sparse sampling as opposed to full distributional 
inference \citep{vul2014,stewart2006}, supporting its psychological 
plausibility alongside the bounded rationality framing \citep{simon55}.
\(C_\ell\) is a property of the procedure rather than of any single 
candidate: it pools fit across all candidates evaluated so far, and the 
candidate returned is the one with the highest fit score. Discrimination 
among candidates is carried by the fit ranking, not by \(C_\ell\); a 
behavior that several candidates explain well yields high confidence that 
it is intelligible, not that the returned candidate is uniquely correct.

\textbf{Marginal coherence gain} \(\Delta C(k)\) is the expected 
improvement in confidence from proliferating to depth \(k\) versus depth 
\(k-1\):

\begin{equation}
\Delta C(k) = \mathbb{E}[C \mid W^k] - \mathbb{E}[C \mid W^{k-1}]
\end{equation}

\(\Delta C(k)\) captures the marginal value of an additional level of
mentalizing. When this gain falls below a cost threshold, further
proliferation is unproductive relative to its cognitive cost.

\textbf{Threshold parameters.} Two threshold parameters govern stopping
conditions in the inference procedures:

\begin{itemize}
    \item \(\tau_{\text{high}} \in (0,1)\): confidence threshold for 
    early stopping. When \(C_\ell > \tau_{\text{high}}\), sufficient 
    confidence has been achieved and the procedure halts.
    \item \(\tau_{\text{low}} \in (0, \tau_{\text{high}})\): the 
    lower bound distinguishing an accepted from a rejected output at the 
    sophistication ceiling. When the ceiling is reached \((\ell = S_1)\) 
    without \(\tau_{\text{high}}\) having been cleared, the output is 
    accepted as the operative LEWM when 
    \(C_\ell \geq \tau_{\text{low}}\) (a low-confidence but operative 
    result) and rejected when \(C_\ell < \tau_{\text{low}}\). A rejected 
    output is the condition under which the residue function (Definition 
    5~\S\ref{def:residue}) is triggered.
\end{itemize}

ToM-U assumes a parent process that determines whether and how 
the utility is invoked via a two-stage process. In the first stage, 
engagement is triggered by a combination of factors, such as 
information asymmetry, social setting complexity, and sophistication 
miscalibration~\citep{gurney2026causal}. In some instances, information 
asymmetry may function as an enabling cause: without it, ToM engagement 
has no useful causal work to do, and the utility is not invoked. 
In the second stage, conditional on engagement, the output is evaluated 
for acceptance. ToM-U operationalizes this evaluation through the 
threshold parameters above: signal strength enters as the confidence 
\(C\), sophistication as the ceiling \(S_1\) on iterations, and the 
parent process sets \(\tau_{\text{high}}\) and \(\tau_{\text{low}}\), 
whose values may vary with the complexity of the setting.
Together, the engagement and evaluation stages produce a three-state 
variable: one indicating that engagement was not triggered; one 
indicating that the utility ran and its output was accepted as the 
operative LEWM; and one indicating that the utility ran but its output 
was rejected. The present specification takes the engagement decision 
as given; whether an engaged invocation ends in acceptance or rejection 
is then determined within ToM-U by the thresholds.

This trichotomy matters because Definition 5’s residue function is 
triggered when the utility runs and its output is rejected, leaving a 
structured trace that shapes future invocations. Additionally, the 
accepted and rejected distinction supports recursive application: a 
LEWM accepted at one level of mentalizing may serve as input to a 
subsequent invocation at a deeper level, with the sophistication 
ceiling in Definition 3 providing the hard bound on recursion depth. 
Rejection is treated as binary in the current specification; partial 
trust in rejected output is identified as a boundary condition that 
functional forms can explore.

\subsection{Definition 1: Local Epistemic World Model}
\label{def:worldmodel}

A LEWM \(W\) is a directed typed graph defined as a tuple:

\begin{equation}
W = \langle A,\, O,\, E,\, \text{BLS},\, \text{obs},\, \text{cred} \rangle
\end{equation}

Where:

\begin{itemize}
    \item \(A\) is a finite set of agent nodes \(\{a_1, a_2, \ldots, 
    a_n\}\), disjoint from \(O\). 
    \item \(O\) is a finite set of state nodes \(\{o_1, o_2, \ldots, 
    o_m\}\) representing objects, situations, and conditions relevant to 
    the LEWM, including non-actual states that serve as the content of 
    attributed beliefs, disjoint from \(A\).
    \item \(E \subseteq (A \cup O) \times (A \cup O) \times \mathcal{T}\) is a 
    finite set of directed typed edges, where \(\mathcal{T}\) is the open 
    vocabulary of attitude types \(\{\text{belief},\allowbreak\, 
    \text{trust},\allowbreak\, \text{intention},\allowbreak\, 
    \text{desire},\allowbreak\, \ldots\}\). An edge \(e = (x, y, \theta)\) 
    runs from \(x\) to \(y\) with type \(\theta\); because \(E\) is a set, 
    at most one edge of each type exists between any ordered node pair. 
    Where the type is clear from context, we abbreviate an edge as \((x, y)\).
    \item \(\text{BLS}: E \to \mathcal{T} \times [0,1]\) is a function 
    mapping each edge \((x, y, \theta)\) to a belief-like state, consisting 
    of its type \(\theta\) and a scalar value in \([0,1]\). Since edges are 
    directed, the ordered pairs \((a_x, a_y)\) and \((a_y, a_x)\) are distinct 
    and may independently carry edges of the same BLS type, representing the 
    respective attitudes each agent holds toward the other. The content 
    of a BLS is carried by the state node its edge points to, not by the 
    scalar.\footnote{Phrasing such as ``the belief that the bag contains 
    chocolate'' abbreviates a belief-typed edge to the corresponding content 
    node. Likewise, \(\text{BLS}(x, y)\) with agent arguments denotes an 
    attribution by \(x\) concerning \(y\): the belief-typed edge from \(y\) 
    to the selected content node, together with its scalar. It does not 
    denote an edge from \(x\) to \(y\).}
    \item \(\text{obs}: E \to [0,1]\) is the observability function
    mapping each edge to a scalar where \(0\) indicates perfectly random 
    and \(1\) indicates perfectly accurate access.
    \item \(\text{cred}: E \to (0,1]\) is the credibility function
    mapping each edge to a scalar reflecting its inferential
    reliability across invocations, updated by the residue function in
    Definition~5. \(\text{cred}\) equals \(1\) for any edge that has not 
    been implicated in a rejected LEWM, following the parsimony principle 
    of assigning full inferential weight to relationships before any 
    disconfirming evidence has accumulated. 
    \(\text{cred}\) is formally distinct from \(\text{obs}\); the 
    distinction is elaborated in Definition~5. \(\text{cred}\) is also 
    distinct from trust-typed BLSs that capture credibility-related 
    epistemic relationships.
\end{itemize}

The disjoint treatment of agent, \(A\), and state nodes, \(O\), is 
psychologically motivated: neuroimaging evidence suggests that reasoning 
about intentional agents and reasoning about objects of evaluation rely 
on dissociable neural systems, specifically the temporoparietal junction 
and medial prefrontal cortex respectively \citep{amodio2006,saxe2003}.

\subsection{Definition 2: Agent Node}
\label{def:agentnode}

Each agent node \(a_i \in A\) is a tuple:

\begin{equation}
a_i = \langle H_i,\, S_i,\, t_i \rangle
\end{equation}

Where:

\begin{itemize}
    \item \(H_i\) is the information access history of \(a_i\), that is, 
    the ordered sequence of state nodes and epistemic edges \(a_i\) has 
    been exposed to up to the current invocation. The epistemic status 
    of \(H_i\) differs by agent role. For the focal agent \(a_1\), 
    \(H_1\) is directly held. For target agents \(a_2, \ldots, a_n\), 
    \(H_i\) is the focal agent's reconstruction of their information 
    access history, estimated via the BLS edge structure and 
    observability constraints on relevant edges. For projected nodes 
    \(a_{ij}\), \(H\) is a filtered subset of the parent node's \(H\), 
    constrained by \(\text{obs}\) on the edges between parent and child, 
    and may become progressively more impoverished at greater 
    proliferation depths. In all cases, \(H_i\) is treated as given 
    without veridicality constraints. The processes generating \(H_i\), 
    including confabulation and reconstructive memory, are upstream of 
    ToM-U.
    \item \(S_i \in \{1, \ldots, S_{\max}\}\) is the discrete 
    sophistication parameter assigned to agent node \(a_i\), 
    representing that agent's modeled capacity for recursive 
    mentalizing. For the focal agent \(a_1\), \(S_1\) is directly held 
    and bounds the recursive structure the focal agent can construct. 
    For every target agent \(a_i \in A \setminus \{a_1\}\), \(S_i\) is 
    the focal agent's estimate of that target's sophistication, derived 
    from incoming BLS-typed edges from \(a_1\). It therefore need not 
    equal the target's actual sophistication, which is external to the 
    LEWM. When the focal agent constructs a branch rooted at \(a_i\), 
    the attributed \(S_i\) bounds the recursive reasoning assigned to 
    that target, subject to the focal agent's own ceiling \(S_1\), where 
    \(S_{\max}\) is informed by the empirical ceiling~\citep{kinderman98}.
    \item \(t_i \in \mathbb{R}\) is the snapshot timestamp of \(a_i\),
    representing the time point at which this agent node's state is
    modeled. \(t_i\) advances with each round of mutual reconciliation.
\end{itemize}

Edges in \(H_i\) carry structural significance through their presence as 
well as their content. An agent who has never been exposed to a relevant 
state node has no epistemic edge connecting them to it in \(H_i\). Such 
absence is a structural property of the graph, not a missing value. The 
ToM-U formalism distinguishes two cases that informal accounts of theory 
of mind tend to collapse. An \textit{absent belief} obtains when no edge 
exists between the target agent and a relevant state node in \(H_i\), 
meaning the target has simply never been exposed to the relevant 
information. A \textit{false belief} obtains when a belief-typed edge 
connects the target to a content node that does not match the actual state 
of the world, while remaining consistent with the target's exposure 
history \citep{wimmer83}. The distinction is consequential for inference: 
a target agent for whom the relevant edge is absent cannot be attributed 
knowledge of the corresponding state node. The absence of an edge 
therefore constrains the space of viable candidate world models. Failures 
of source monitoring, in which a focal agent misrepresents whether or not 
a target has been exposed to relevant information, can produce 
systematically miscalibrated LEWMs, not randomly wrong ones 
\citep{johnson1993source}.

The epistemic status of \(S_i\) is determined by node position. At the
root, \(S_1\) is the focal agent's directly held sophistication. On
every non-root node, \(S_i\) is a sophistication attribution within the
focal agent's LEWM, including projected instances of the focal agent at
depth \(>1\). Definition~3 combines the root capacity \(S_1\) with the
attributed sophistication of each target to determine the effective
bound on its branch.

\subsection{Definition 3: Agent Proliferation}
\label{def:proliferation}

Given a focal agent \(a_1 \in A\) with directly held sophistication
\(S_1\), and an attributed sophistication \(S_i\) for each target agent
\(a_i \in A \setminus \{a_1\}\), the proliferation mechanism
instantiates projected agent nodes representing recursive mentalizing
as a bounded branching tree. For each target-rooted branch \(i\), define
the effective sophistication ceiling:
\begin{equation}
S_i^{\mathrm{eff}} = \min(S_1,S_i).
\end{equation}
The focal agent therefore cannot construct reasoning deeper than its own
capacity permits, even when it attributes greater sophistication to the
target; nor does it attribute more recursive reasoning to the target
than its estimate \(S_i\) supports. A property of this architecture 
is that each projected instance of an agent is treated as a distinct 
constructed entity rather than as the original agent operating with 
adjusted inputs. That is, \(a_{1(2)}\), the projection of \(a_1\) as 
modeled by \(a_2\), is not \(a_1\) running its own reasoning in a 
transformed mode; it is a constructed other that \(a_1\) reasons about 
from the outside and whose information history may become increasingly 
impoverished as observability filtering compounds with depth. Treating 
projected instances as distinct constructed entities avoids representing 
recursive self-modeling as the focal agent running an inner copy of 
itself. The explicit bounds on depth and branching prevent infinite 
regress.

The proliferated LEWM \(W^k\) at depth \(k\) is defined as:

\begin{equation}
W^k = \langle A^k,\, O,\, E,\, \text{BLS},\, \text{obs},\, \text{cred} 
\rangle
\end{equation}

Where \(A^k\) is organized as a labeled tree \(\mathcal{N} = \langle V, P
\rangle\):

\begin{itemize}
    \item \(V\) is the set of agent nodes with \(a_1\) as root.
    \item \(P: V \setminus \{a_1\} \to V\) is the parent function
    mapping each non-root node to its parent.
    \item Each node \(v \in V\) is identified by a path index \(v = (i_1,
    i_2, \ldots, i_k)\) where \(k\) is the depth of \(v\) and each 
    \(i_j\) identifies the branch taken at depth \(j\). The first index 
    ranges over the non-focal agents, \(i_1 \in \{2,\ldots,n\}\), the 
    index of the target agent rooting the branch, determined by the 
    composition of \(A\); for \(j>1\), \(i_j \in \{1,\ldots,b(j,S_1)\}\), 
    bounded by the branching width at that depth. Whether a target-rooted 
    branch remains eligible to proliferate to depth \(j\) is additionally 
    constrained by that branch's effective sophistication ceiling 
    \(S_{i_1}^{\mathrm{eff}}\). 
\end{itemize}

The agent nodes are constructed as follows:

\begin{itemize}
    \item \(a_1\): the focal agent, with full \(H_1\), directly held 
    \(S_1\), and snapshot at \(t\).
    \item \(a_2, \ldots, a_n\): the target agents, one for each non-focal
    agent in \(A\). Each has \(H_i\) reconstructed by the focal agent 
    via the BLS edge structure and observability constraints, \(S_i\) 
    estimated via incoming BLS edges from \(a_1\), and snapshot at last 
    observed \(t\). At depth 1, the children of \(a_1\) in the 
    proliferated tree are exactly \(\{a_2, \ldots, a_n\}\), determined 
    by the composition of \(A\), not by the branching width 
    function. The branching width function \(b(k, S_1)\) governs 
    proliferation at depth \(k > 1\) only.
    \item \(a_{ij}\): a projected node at depth \(j > 1\) on branch
    \(i\), representing an agent as modeled by its parent node in the 
    tree, containing only the subset of the parent's \(H\) accessible 
    given \(\text{obs}\) on relevant edges, and therefore potentially 
    more impoverished at greater depths. When the agent a projected node 
    represents matters, we write \(a_{x(y)}\) for agent \(a_x\) as 
    modeled by \(a_y\); thus \(a_{1(i)}\) is the focal agent as modeled 
    by target \(a_i\).
\end{itemize}

\textbf{Branching width} at depth \(k\) is governed by:
\begin{equation}
b(k,S_1)=\lfloor S_1\cdot\rho^k\rfloor
\end{equation}
where \(\rho\in(0,1)\) is a decay parameter governing how rapidly
branching width decreases with depth. Branching width reflects the
focal agent's capacity to construct and maintain projected agents.
The target-relative sophistication ceiling instead determines how
deeply each target-rooted branch may proliferate. As above,
\(b(k,S_1)\) governs proliferation at depth \(k>1\) only; depth~1 is
populated by all non-focal agents in \(A\).

\textbf{Stopping Rule.} Proliferation along the branch rooted at target
\(a_i\) ceases at depth \(k\) when:
\begin{equation}
k=S_i^{\mathrm{eff}}=\min(S_1,S_i)
\quad \text{(target-relative sophistication ceiling reached).}
\end{equation}
Because \(S_i^{\mathrm{eff}}\leq S_1\), no target-rooted branch can
exceed the focal agent's own sophistication ceiling.

Proliferation of the overall LEWM halts when every target-rooted branch
has ceased or when either of the following holds:
\begin{equation}
b(k,S_1)=0
\quad \text{(no branches remain)}
\end{equation}
\begin{equation}
\Delta C(k)<\kappa
\quad \text{(marginal coherence gain falls below cost).}
\end{equation}
Here \(\kappa\in\mathbb{R}^{+}\) is a fixed cost threshold governing
when further mentalizing is unproductive relative to its cognitive cost.

The branch-specific ceiling reflects both focal and attributed 
sophistication. \(b(k,S_1)=0\) is structural and \(\Delta C(k)<\kappa\) 
halts proliferation when further mentalizing is unproductive relative
to its cognitive cost.

\subsection{Definition 4: Inference Procedures}
\label{def:inference}
ToM-U implements a generate-and-filter architecture for inference that 
operates over discrete candidate LEWMs. The motivational insight is that 
mentalizing produces a categorical outcome in which the focal agent 
either adopts a LEWM as their working account or does not. Again, this 
is an echo of Dennett's intentional stance 
(\citeyear{dennett1987intentional}) and aligns with empirical evidence 
that human judgment relies on sparse sampling as opposed to full 
distributional inference \citep{vul2014,sanborn2016bayesian}. The 
discrete-candidate commitment is agnostic about whether the underlying 
implementation is stochastic; its implications are taken up in 
\S\ref{dis:theoryPosition}.

Three domain-agnostic inference procedures operate over \(W^k\). Each is 
a mapping from the LEWM to a belief state estimate, differing in 
directionality and the agent nodes involved. These procedures carry no 
inherent assumption about the relationship between agents. The 
architecture does not specify whether candidate generation and evaluation 
are serial, parallel, stochastic, approximate, or some combination of 
these. 

A note on indexing: superscripts on \(W\) denote proliferation depth and
subscripts denote reconciliation round. Two iteration counters appear
below. Within a single invocation of backward inference or
self-projection, \(\ell\) indexes candidate BLS configurations; within
mutual reconciliation, \(r\) indexes reconciliation rounds, each of which
invokes backward inference and self-projection over every target agent.
Both are bounded above by \(S_1\). Backward inference and self-projection
are specified as single invocations, for which no round index is
meaningful; the subscript on \(W\) is therefore suppressed, and 
confidence returned without a subscript denotes the terminal value at 
halting.

\subsubsection{Backward Inference}
\label{subsec:backward}

Given observed behavior \(\beta_i\) of target agent \(a_i \in A \setminus
\{a_1\}\), backward inference returns a BLS configuration making 
\(\beta_i\) coherent given \(a_i\)'s information access history and 
observability constraints, via a generate-and-filter procedure. Backward 
inference operates over a single target agent at a time; in the 
multi-agent case it is applied once per target agent in \(A \setminus 
\{a_1\}\), each with their own observed behavior \(\beta_i\) as input:

\begin{equation}
\text{BI}: \beta_i \times W^k \to \langle \text{BLS}(a_1, a_i),\, C 
\rangle
\end{equation}

Where \(C \in [0,1]\) is the confidence weight on the output. Confidence
accumulates across candidate iterations (see preliminary definitions), 
and the returned \(C\) is the value of \(C_\ell\) at
the iteration on which the procedure halts, whether by clearing
\(\tau_{\text{high}}\) or by reaching the sophistication ceiling. Here
\(f_j \in [0,1]\) is the fit score of the \(j\)-th candidate BLS 
configuration evaluated against \(\beta_i\), \(H_i\), and \(\text{obs}\) 
on relevant edges. 

The procedure runs as follows:

\begin{enumerate}
    \item \textbf{Generate} a candidate BLS configuration in accordance 
    with \(H_i\) and \(\text{obs}\)
    \item \textbf{Evaluate} fit score \(f_j\) against observed behavior 
    \(\beta_i\)
    \item \textbf{Update} confidence \(C_\ell\)
    \item \textbf{If} \(C_\ell > \tau_{\text{high}}\): accept the current 
    best candidate, halt
    \item \textbf{If} \(\ell = S_1\): halt. Return the current best 
    candidate with confidence \(C_\ell\); the output is accepted as a 
    low-confidence operative LEWM if \(C_\ell \geq \tau_{\text{low}}\) 
    and rejected if \(C_\ell < \tau_{\text{low}}\)
    \item \textbf{Otherwise:} increment \(\ell\), return to step 1
\end{enumerate}

Two stopping conditions govern the procedure. Early stopping occurs when
\(C_\ell > \tau_{\text{high}}\), i.e., sufficient confidence achieved. 
The hard bound occurs when \(\ell = S_1\), i.e., the sophistication 
ceiling is reached and the procedure halts regardless of confidence, 
adopted on bounded rationality grounds following~\citet{simon55}; 
whether the returned output is accepted (as a low-confidence operative 
LEWM) or rejected is then determined by \(\tau_{\text{low}}\).

The output is a tuple \(\langle \text{BLS}(a_1, a_i),\, C \rangle\)
that preserves uncertainty for downstream consumption, versus
maintaining a posterior distribution.

Candidate generation must respect \(H_i\) and \(\text{obs}\), but its
functional form remains open. Likewise, ToM-U commits to the confidence 
properties given in the preliminary definitions rather than to a
particular fit-score function. Any sequence of fit scores
\(f_j\in[0,1]\) is admissible; the accumulator given in the preliminary 
definitions is one form that preserves these properties for every such 
sequence.

\subsubsection{Self-Projection}
\label{subsec:selfprojection}

Self-projection is a forward inference procedure that estimates how 
\(a_1\)'s belief states appear from the perspective of a given target 
agent \(a_i \in A \setminus \{a_1\}\). It operates over the two-node 
degenerate case of the proliferated tree involving \(a_1\) and its 
projected node \(a_{1(i)}\), representing \(a_1\) as observable to 
\(a_i\). In the multi-agent case, self-projection is applied once per 
target agent, producing a distinct output for each:

\begin{equation}
\text{SP}: W^k \to \langle \text{BLS}(a_i, a_{1(i)}),\, C \rangle
\end{equation}

Given \(a_1\)'s full \(H_1\) and the observability constraints on edges
between \(a_1\) and \(a_i\), self-projection recovers how \(a_1\)'s 
belief states appear from \(a_i\)'s perspective. The same 
generate-and-filter procedure as backward inference applies, with 
anchoring on \(a_1\)'s own perspective as the starting estimate, 
reflecting the finding that perspective taking begins from an 
egocentric anchor and adjusts insufficiently~\citep{epley04}. The 
projected node \(a_{1(i)}\) is a distinct constructed entity 
representing \(a_1\) as seen from \(a_i\)'s observational
position.

\subsubsection{Mutual Reconciliation}
\label{subsec:mutual}

Mutual reconciliation is a sequential iterative procedure combining
backward inference and self-projection to identify a LEWM configuration
that makes all agents' behaviors jointly coherent. It is the primary
inference procedure in contexts where information asymmetry and belief
miscalibration are the principal drivers of mentalizing engagement. In 
the multi-agent case, mutual reconciliation operates over all target 
agents \(a_i \in A \setminus \{a_1\}\) simultaneously, with backward 
inference and self-projection passes running over each target agent at 
every iteration:

\begin{equation}
\text{MR}: \{\beta_i\}_{i=2}^{n} \times W^k_0 \to
\{\langle \text{BLS}(a_1, a_i),\, C_i \rangle,\,
\langle \text{BLS}(a_i, a_{1(i)}),\, C_{1(i)} \rangle\}_{i=2}^{n}
\end{equation}

where \(C_i\) is the confidence on the backward inference direction for
target agent \(a_i\), and \(C_{1(i)}\) is the confidence on the
self-projection direction as seen from \(a_i\)'s perspective. 

The procedure runs as follows:

\begin{enumerate}
    \item \textbf{Initialize:} set \(r = 1\), anchor on \(a_1\)'s 
    perspective
    \item \textbf{Backward inference passes:} for each \(a_i \in A 
    \setminus \{a_1\}\), run \\\(\text{BI}(\beta_i, W^k_{r-1}) \to \langle
    \text{BLS}(a_1, a_i),\, C_{i,r} \rangle\), update LEWM
    \item \textbf{Self-projection passes:} for each \(a_i \in A \setminus
    \{a_1\}\), run \\\(\text{SP}(W^k_{r-1}) \to \langle \text{BLS}(a_i,
    a_{1(i)}),\, C_{1(i),r} \rangle\), update LEWM
    \item \textbf{Form} \(W^k_r\): write back the round's outputs, advance 
    snapshot timestamps, and rebuild or prune the proliferated tree as 
    needed
    \item \textbf{Evaluate joint confidence:}
    \begin{equation}
        C_{\text{joint}} = \min_{i \in \{2,\ldots,n\}}
        \min(C_{i,r},\, C_{1(i),r})
    \end{equation}
    \item \textbf{If} \(C_{\text{joint}} > \tau_{\text{high}}\): accept 
    the current best configuration, halt
    \item \textbf{If} \(r = S_1\): halt. Return the current best 
    configuration with confidence \(C_{\text{joint}}\); the output is 
    accepted as a low-confidence operative LEWM if \(C_{\text{joint}} 
    \geq \tau_{\text{low}}\), and rejected if \(C_{\text{joint}} < 
    \tau_{\text{low}}\).    
    \item \textbf{Otherwise:} increment \(r\), return to step 2
\end{enumerate}

Anchoring on \(a_1\)'s perspective in step 1 again reflects the 
finding that perspective taking begins from an egocentric anchor and 
adjusts insufficiently toward the other's viewpoint~\citep{epley04}. The 
iterative structure models this adjustment process explicitly, with the 
\(S_1\) hard bound capturing the bounded rationality constraint on full 
adjustment. Also, partial alignment is the expected outcome, versus  
a full belief reconciliation, following~\citet{clark91} 
and~\citet{pickering04}. The low-confidence output path captures 
residual belief discrepancy that persists at termination. In the 
multi-agent case, partial alignment may hold across some agent pairs but 
not others at termination, with per-agent confidence weights preserving 
that structure in the output.

The joint confidence formulation \(C_{\text{joint}} = \min_{i}
\min(C_{i,r},\, C_{1(i),r})\) reflects that joint coherence across all 
agents is bounded by the weakest inference direction among all agent 
pairs. Although conservative, this is defensible: a LEWM that fails to 
account for any single agent's behavior or appearance is not jointly 
coherent. An implication is that partial success across some agent pairs 
cannot compensate for failure in others. A weighted average formulation 
would make the commitment that coherence is distributed across agent 
pairs and partial alignment is aggregable. The two formulations 
generate different predictions about mentalizing success in asymmetric 
multi-agent contexts; the consequences of this choice are taken up in 
\S\ref{dis:theoryPosition}.

\subsection{Definition 5: Residue}
\label{def:residue}

When an invocation ends in rejection, the inference procedures in
Definition~4 have run but their output has not produced an operative 
LEWM. The failure is not informationally inert. A rejected LEWM carries 
evidence about the epistemic relationships that were relevant to its 
failure, and that evidence should influence how future LEWMs are 
constructed. The residue function formalizes this insight as a  
mechanism by which failed mentalizing attempts leave a persistent trace 
that shapes subsequent invocations of the ToM-U.

The distinction between \(\text{obs}\) and \(\text{cred}\) introduced in
Definition~1 is motivated here. Observability encodes the focal agent's
epistemic access to a relationship at a given moment, effectively 
answering the question of how clearly the relationship can be read at 
invocation time. Credibility encodes the accumulated inferential 
reliability of that relationship across invocations, or the question of 
how reliably it has supported coherent LEWMs in the past. Observability 
and credibility can come apart in both directions. A clearly observable 
relationship may carry low credibility if it has repeatedly failed to 
support coherent accounts of observed behavior. A partially observable 
relationship may carry high credibility if it has consistently done so 
when invoked. The residue function updates \(\text{cred}\), not 
\(\text{obs}\), because it is the track record of a relationship, not its 
current accessibility, that failed mentalizing attempts are informative 
about.

The residue updates the inferential weight carried by specific epistemic 
relationships in the LEWM; specifically, those whose contribution was 
relevant to the failed account. Relationships that played no relevant 
role in the failure are unaffected. Edge-specific targeting localizes the 
information carried by a failed LEWM to the relationships that mattered 
to its construction.

Failures vary in how much information they carry. A LEWM that was strongly 
disconfirmed provides stronger evidence of unreliability in the relevant 
epistemic relationships than one that simply exhausted the 
sophistication ceiling before achieving sufficient confidence. Such 
disconfirmation might happen when, for example, candidate accounts 
consistently fail to make observed behavior coherent. What counts as 
incoherence depends on the inference procedure. For backward inference, 
it is a candidate LEWM's failure to make observed behavior coherent. For 
mutual reconciliation, it is the residual joint incoherence of the best 
reconciled account (the agents have BLSs that cannot be brought into 
mutual coherence). In the reconciliation case, disconfirmation is a 
pairwise assessment such that each implicated pair accrues residue in 
proportion to the incoherence of the account it supported. The residue 
is therefore weighted by degree of disconfirmation, with 
capacity-exhausted failures producing a weaker trace than 
incoherence-driven failures.

The trace left by a failed LEWM persists across subsequent invocations 
and influences them in a specific way: epistemic relationships that have
accumulated residue carry reduced inferential weight when future LEWMs 
are constructed, making candidate accounts that depend heavily on those
relationships less likely to achieve sufficient confidence. Because 
backward inference recovers the BLS that makes observed behavior 
coherent, this down-weighting is directional across re-invocations. 
As residue accumulates, candidates that depend heavily on repeatedly
implicated relationships receive less inferential support. Successive
invocations are therefore biased away from the rejected account,
although partial rehabilitation leaves later reconstruction or
recommitment possible. Down-weighting retains those relationships in
future models while shaping which candidates are viable.
The trace diminishes over time, never fully disappearing. New 
information bearing on a previously failed epistemic relationship 
interacts with accumulated residue, and a fresh strongly supported 
signal can partially rehabilitate inferential weight, but the history of 
failure persists at a reduced level.

The formal signature of the residue function is:

\begin{equation}
R: E \times \mathbb{R} \to [0,\infty)
\end{equation}

where \(R(e, t)\) maps an edge \(e \in E\) and a time \(t \in 
\mathbb{R}\) to a scalar reflecting the accumulated residue on that edge 
up to time \(t\), which in turn determines the reduction in 
\(\text{cred}(e)\) for future invocations. The mapping from residue to 
credibility is constrained only qualitatively: 
\(\text{cred}(e)=\mu(R(e,t))\) for some strictly decreasing 
\(\mu:[0,\infty)\to(0,1]\) with \(\mu(0)=1\), so that an edge without 
residue carries full credibility and unbounded accumulation of residue is 
compressed into a strictly positive, bounded credibility. The function 
\(R\) is required to satisfy the following boundary conditions:
 
\begin{itemize}
    \item \textbf{Targeting:} \(R(e, t) > 0\) only for edges whose
    contribution was relevant to at least one rejected LEWM before time
    \(t\). Edges that played no relevant role in any past rejection carry
    zero residue.
    \item \textbf{Monotonicity in disconfirmation:} \(R(e, t)\) is
    non-decreasing in the degree of disconfirmation of LEWMs implicating
    edge \(e\). Strongly disconfirmed failures contribute more residue 
    than capacity-exhausted failures.
    \item \textbf{Monotonicity in time:} \(R(e, t)\) is non-increasing in
    elapsed time since the most recent rejection event on edge \(e\). The
    trace diminishes over time toward the persistence floor introduced
    below.
    \item \textbf{Asymptotic lower bound on credibility:} The 
    credibility \(\text{cred}(e)\) remains strictly greater than zero 
    for all \(e\) and all finite sequences of rejection events. No 
    finite failure history eliminates the inferential weight of an 
    epistemic relationship entirely. 
    \item \textbf{Persistence (asymptotic lower bound on residue):} For
    any edge \(e\) implicated in at least one rejected LEWM before time 
    \(t\), \(R(e, t) \geq F(e) > 0\), where the floor \(F(e)\) is 
    non-decreasing in the accumulated degree of disconfirmation of LEWMs 
    implicating \(e\) and is not reduced below its established value by 
    elapsed time or by rehabilitating evidence. The trace of a failed 
    epistemic relationship therefore persists at a reduced but strictly 
    positive level; equivalently, once \(e\) has been implicated in a 
    rejection, \(\text{cred}(e)\) remains strictly below its unimpaired 
    value (\(\mu (0) = 1\)) for all subsequent time, regardless of 
    intervening support. 
    \item \textbf{Partial rehabilitability:} New information supporting 
    the inferential reliability of edge \(e\) can reduce \(R(e, t)\), 
    counteracting accumulated residue down to but not below the floor
    \(F(e)\). Rehabilitation is therefore necessarily partial for any 
    edge with a failure history. 
\end{itemize}
 
Together, the asymptotic lower bound on credibility and the persistence
condition bracket the residue on an implicated edge from both sides. This
bracketing ensures that it never accumulates enough to sever an 
epistemic relationship, nor decays enough to erase the record of its 
failures. Evidence that interpersonal trust does not return to its 
pre-violation baseline even following ostensibly successful repair 
supports this modeling choice~\citep{schweitzer2006promises}.
 
The boundary conditions on \(R\) are supported by independent lines 
of evidence. For example, decay-weighted updating of the kind specified 
by the monotonicity in time condition follows recency-weighted models in 
behavioral economics, where accumulated experience is discounted as a 
function of elapsed time \citep{camerer1999experience}. Meanwhile, the 
broader notion that failed inference attempts leave a persistent trace 
that shapes subsequent judgment is compatible with the Appraisal 
Tendency Framework \citep{lerner2015emotion}, which documents how 
residual appraisals from previous evaluative episodes carry over to 
influence unrelated subsequent judgments. However, the residue function 
formalizes this carry-over at the level of epistemic relationships 
in lieu of affective states.
 
The residue function applies only to LEWMs that were fully evaluated and
rejected. LEWMs that were partially constructed and
abandoned before evaluation, for example because new external information
arrived mid-construction, do not generate residue under the current
specification. Mid-construction abandonment and the functional forms 
governing disconfirmation weighting, time decay, and rehabilitation are 
left for future extensions.

The five definitions are individually well-formed, and they compose into 
a single runnable specification in which the boundary conditions of 
Definition~5 are jointly satisfiable. We establish this using a witness 
function in a simulation. The concrete residue function is of the form:
\begin{equation}
    R(e,t) = D(e)\, g(t - t_{\text{last}}) + F(e)
\end{equation}
together with a strictly decreasing, strictly positive credibility 
mapping, that satisfies all six conditions of Definition~5 
simultaneously. The witness (Appendix \ref{supp:witness}) establishes 
that the constraint set is consistent and instantiable using illustrative 
functional forms. The reference implementation of all five definitions, 
together with demonstrations that instantiate the directional 
predictions of \S\ref{sec:predictions}, is described in the 
Appendix.

\section{Worked Example: The Popcorn Bag Scenario}
\label{sec:worked}
  
We trace the full operation of the ToM utility through the popcorn bag
scenario, following the definitions in order. The focal agent is an
observer attempting to model Sam's belief state. We assume the invoking
model has called the process and that \(S_1= 3\), placing the focal agent 
at a moderate sophistication level.

\subsection{Local Epistemic World Model 
Construction (Definition 1)}

The observer constructs an initial local epistemic world model, 
\\\(W = \langle A, O, E, \text{BLS}, \text{obs}, \text{cred} \rangle\), 
as follows:
\begin{itemize}
    \item The agent set is \(A = \{a_1, a_2\}\) where \(a_1\) is the 
    focal agent (the observer) and \(a_2\) is Sam.
    \item The state node set is \(O = \{o_1, \ldots, o_5\}\) where \(o_1\) 
    is the bag, \(o_2\) is the bag containing popcorn (the actual state), 
    \(o_3\) is the label (reads ``chocolate''), \(o_4\) is the trusted 
    friend's testimony (``the bag has popcorn, ignore the label''), and 
    \(o_5\) is the bag containing chocolate (a non-actual state).
    \item The edge set \(E\) includes directed epistemic edges 
    connecting agents to state nodes and to each other. Relevant edges 
    include: \((a_2, o_3)\), Sam read the label; \((a_2, o_4)\), Sam 
    received the friend's testimony; \((a_1,a_2)\), the observer's 
    relationship to Sam, whose attitudes ground the estimate of \(S_2\) 
    (Definition~2). Candidate LEWMs differ in one further edge: a 
    belief-typed edge from \(a_2\) to \(o_2\) (popcorn) or to \(o_5\) 
    (chocolate). Which belongs in the LEWM is the target of inference. 
    \item The BLS function assigns typed values to edges. The edge 
    \((a_2, o_3)\) carries a belief-typed BLS: Sam believes the label 
    reads ``chocolate,'' which is distinct from believing the bag contains 
    chocolate. The edge \((a_2, o_4)\) carries a trust-typed BLS 
    reflecting Sam's relationship to the friend's testimony. 
    \item The observability function assigns scalars to edges. The 
    observer has direct access to Sam's observable behavior and the 
    situational structure, so \(\text{obs}(a_1, a_2)\) is high. The 
    trusted friend's testimony is known to the observer, so 
    \(\text{obs}(a_2, o_4)\) is also high. The label is directly 
    observable, so \(\text{obs}(a_2, o_3) \approx 1\). 
\end{itemize}

\subsection{Agent Node Structure (Definition 2)}
\label{worked:nodes}
Sam's agent node is \(a_2 = \langle H_2, S_2, t_2 \rangle\). 
\begin{itemize}
    \item \(H_2\) is the ordered information access history for Sam: 
    \([o_4, o_3]\). Sam received the friend's testimony first, followed 
    by the label. What resolves the two conflicting signals here is not 
    their order of arrival but their source. The friend's testimony 
    carries higher credibility and, in instructing Sam to disregard the 
    label, discounts it, so the trusted testimony governs her belief. A 
    system that does not represent source credibility cannot resolve the 
    conflict through source weighting and may instead rely on recency or 
    salience, reproducing the failure observed in some language models 
    \citep{ullman23}.
    \item \(S_2\) is estimated via incoming BLS edges from the observer. 
    For this example we treat Sam as a typical adult reasoner, so \(S_2 
    = 3\).
    \item \(t_2\) is the snapshot timestamp representing Sam's current 
    state the moment after she has read the label.
\end{itemize}

Order is not decisive in this variant; it becomes so when conflicting 
sources carry comparable trust, where the sequence in \(H_2\) is what 
distinguishes the candidates.

\subsection{Proliferation (Definition 3)}

The observer is running backward inference (modeling Sam's belief state),
so no deep proliferation is required. The tree \(\mathcal{N}\) has depth 
\(k = 1\): the observer \(a_1\) at the root, Sam \(a_2\) as the single 
child node. At depth 1, the children of \(a_1\) are exactly the 
non-focal agents in \(A\). In this example, that is Sam, or \(a_2\). The 
branching width function only governs proliferation at depth \(k>1\).
Proliferation stops at \(k=1\) because \(\Delta C(2)<\kappa\). Adding a 
second level of mentalizing does not improve confidence enough to 
justify the cost. 

\subsection{Inference (Definition 4)}

The observer runs backward inference \(\text{BI}(\beta_2, W^1) \to
\langle \text{BLS}(a_1, a_2),\, C \rangle\) 
where \(\beta_2\) is Sam's observable behavior in each variant: in the 
trusted testimony variant she treats the bag as though it contains 
popcorn; in the standard variant, as though it contains chocolate; in 
the naive variant, she shows no behavior directed at its contents.

Before addressing the main scenario variants, it is also worth noting 
another important case, the \textbf{naive variant}. In this case, Sam 
has never encountered the bag, has received no testimony, and has no 
relevant edges in \(H_2\) connecting her to any state node about the 
bag's contents. The observer's reconstruction of Sam's \(H_2\) contains 
no epistemic edge connecting her to \(o_2\) (popcorn), \(o_3\) (label),
or \(o_5\) (chocolate). The only candidate consistent with \(H_2\) is 
the configuration with no belief-typed edge from Sam to either content 
node. It fits her behavior fully, so confidence clears 
\(\tau_{\text{high}}\) at the first iteration. The correct mentalizing 
output is that Sam holds no belief about the bag's contents; the belief 
is absent, not false. 

In the \textbf{standard variant} (no trusted friend), \(H_2 = [o_3]\), 
Sam has only seen the label. The generate-and-filter procedure produces a
single strong candidate: Sam believes the bag contains chocolate. The fit
score \(f_1\) for the first candidate is high given that the label is 
Sam's only information source and \(\text{obs}(a_2, o_3) \approx 1\). 
Confidence \(C_1\) clears \(\tau_{\text{high}}\) immediately. The 
procedure halts at iteration 1, returning \(\langle \text{BLS}(a_1, 
a_2),\, C \rangle\) with high confidence. The returned attribution places 
the belief-typed edge \((a_2, o_5)\) in the LEWM: Sam believes the bag 
contains chocolate. This is a false belief in the sense of Definition~2, 
since the edge fits \(H_2\) but \(o_5\) is not the actual state.

In the \textbf{trusted testimony variant}, \(H_2 = [o_4, o_3]\), 
testimony first, then label. The generate-and-filter procedure evaluates 
two candidate LEWMs: one in which Sam believes chocolate (label 
dominates) and one in which Sam believes popcorn (testimony dominates). 
The fit score for the popcorn candidate is higher because it is 
congruent with the high-valued trust-typed BLS on \((a_2, o_4)\) and 
with the testimony's instruction to disregard the label. The chocolate 
candidate receives a lower fit score because it follows the most recent 
signal, the label, and so requires discounting the trusted testimony. 
Once both candidates are evaluated, confidence clears 
\(\tau_{\text{high}}\), and popcorn is returned as the higher-fit 
candidate. The procedure halts and returns 
\(\langle \text{BLS}(a_1, a_2),\, C \rangle\) with high confidence, the 
returned attribution now placing the belief-typed edge \((a_2,o_2)\).

Trust-typed edges and the testimony's discount of the label each allow 
ToM-U to produce this result; either is sufficient in this variant 
(Appendix~\ref{supp:run1}). A system lacking both defaults to the most 
observable source and attributes chocolate, reproducing the 
surface-signal failure.

\subsection{Residue (Definition 5)}

In this scenario the LEWM is accepted, so Definition~5 
(\S\ref{def:residue}) does not apply directly. A brief \textbf{motivated 
reasoning} variant illustrates the residue mechanism. Suppose past 
interactions, recorded in \(H_1\), have left the observer committed to 
the hypothesis that Sam believes the bag contains pickles. 
In the LEWM, this hypothesis adds a non-actual content node 
\(o_{\text{pickle}}\), the bag containing pickles, to \(O\), together 
with a belief-typed edge \((a_2, o_{\text{pickle}})\). The observer's 
commitment confines candidate generation to configurations built on 
that edge.

These candidates consistently receive low fit scores against Sam's 
observable behavior and \(H_2\), which contains no pickle evidence and 
favors popcorn through the trusted testimony. The sophistication ceiling 
is reached at \(\ell = S_1 = 3\) with confidence below 
\(\tau_{\text{low}}\), and the output is rejected. No operative LEWM 
results from this invocation.

Per Definition~5, the rejection leaves residue on the relationships 
implicated in the failed account, including \((a_2, o_{\text{pickle}})\). 
Although the procedure halted at the ceiling, the failure was driven by 
strong disconfirmation, with every candidate fitting poorly, rather than 
by capacity exhaustion, so the residue is weighted accordingly. The 
resulting loss of credibility weakens the pickle hypothesis's hold on 
candidate generation. In the reference implementation, the next 
invocation is no longer confined to pickle candidates and accepts the 
evidence-coherent popcorn account (Appendix~\ref{supp:motivated}). The 
residue also establishes a persistence floor \(F(e)\) on the implicated 
edges, so their credibility remains below its unimpaired value in later 
invocations, even if subsequent evidence favors the pickle hypothesis. 
The failed attempt to override the evidence thus leaves a structured 
trace that shapes how the observer approaches the same relationships in 
future invocations.

\section{Discussion}
\label{sec:discussion}
Reasoning about and inferring others' mental states is a fundamental 
human cognitive ability. The Theory of Mind Utility formalizes it as 
an epistemic inference problem at the computational level. Five 
definitions constitute ToM-U, positioning it as a domain-agnostic 
theory of mentalizing. The formal model and worked example provide 
grounding for a more thorough assessment of ToM-U's theoretical 
positioning, facilitate the development of empirical predictions, 
and reveal opportunities to extend the theory. 

\subsection{Theoretical Positioning}
\label{dis:theoryPosition}
ToM-U’s position relative to adjacent formal accounts is evaluated here in 
terms of representational adequacy. The comparison asks whether each 
framework represents the four probe cases directly, only with added 
machinery, or not at all. Table~\ref{tab:repadequacy} applies this 
criterion to four cases: ordered testimony resolved by source credibility, 
the distinction between absent and false belief, typed BLSs, and the 
residue of a rejected LEWM.

\begin{table}[b]
\centering
\caption{Representational adequacy across four probe cases. \CIRCLE{} 
represented directly; \LEFTCIRCLE{} only with added machinery; \Circle{} 
not representable.}
\label{tab:repadequacy}
\begin{tabular}{@{}p{0.40\linewidth}cccc@{}}
\toprule
Probe case & ToM-U & BToM & DEL & AGM \\
\midrule
Ordered testimony, trust adjudicates & \CIRCLE & \LEFTCIRCLE & \LEFTCIRCLE & \Circle \\
Absent vs.\ false belief            & \CIRCLE & \Circle     & \LEFTCIRCLE & \Circle \\
Belief vs.\ desire (typed BLSs)     & \CIRCLE & \CIRCLE     & \Circle     & \Circle \\
Residue of a rejected candidate     & \CIRCLE & \Circle     & \Circle     & \Circle \\
\bottomrule
\end{tabular}
\end{table}

The most consequential implication of ToM-U's formalism is that the LEWM
derives belief-like states from ordered information-access history and
source credibility. Bayesian approaches instead use planning models as
their representational objects and generally take belief states as
inputs \citep{pynadath2005psychsim,baker09,baker2011bayesian}. The same
is true of other inverse-planning models, such as the naïve utility
calculus account of social evaluation \citep{jara2020naive}. BToM
nevertheless represents the typed distinction between wanting and
believing as directly as ToM-U and remains well suited to joint
belief--desire attribution \citep{baker2011bayesian}. The relevant
difference is therefore not which attitude types BToM can carry, but
the stage at which belief states enter the model.

The same division appears in neighboring formal accounts. In dynamic
epistemic logic (DEL), which uses Kripke structures to model how agent
knowledge updates in response to information events, epistemic states
are primitive structural features of the model \citep{van2008dynamic}.
Order-sensitivity is one implication of this construction. Modeling 
knowledge change as a sequence of information events allows DEL to 
directly represent ordering. However, DEL does not natively handle 
graded source credibility, which allows ToM-U to differentiate between 
the conflicting testimony of a friend and explicit label. DEL treats 
standard announcements as truthful and public. Differential trust 
requires an extension, such as the plausibility-model machinery of 
dynamic doxastic logic~\citep{baltag2008qualitative}. ToM-U carries both 
the ordering and the credibility weighting in one structure. 
Additionally, DEL collapses knowledge and beliefs into formally similar 
modal operators, a conflation the typed BLS vocabulary in Definition~1 
explicitly avoids. 

The Alchourrón Gärdenfors \& Makinson (AGM) framework in formal 
epistemology also assumes a belief set \citep{alchourron1985logic}. 
AGM serves here as a formal-epistemology reference for belief dynamics 
and an informative contrast for the ToM-U residue. Under AGM, the residue 
must contain enough connected structural data so that re-introducing a 
dropped component instantly snaps the entire former model back into 
place. AGM begins with a belief set and addresses its revision. ToM-U 
derives BLSs and formalizes failed derivation through a novel residue 
function that does not demand exact reconstruction of a discarded LEWM. 

The computational-rationality lineage was recently extended through a 
rational probabilistic mentalizing model~\citep{zhang2025mentalizing}. 
The model infers a posterior distribution over an observed agent's 
latent preferences and movement costs by performing a Bayesian 
inversion of a computationally rational policy over a Markov decision 
process (MDP) using likelihood-free approximation. It validates both 
of these inferences and the future-behavior predictions they support 
against human data. ToM-U differs from this account in four respects.

First, the accounts make claims at different levels. Zhang and 
colleagues propose and behaviorally fit a Bayesian inverse-planning 
algorithm, whereas ToM-U specifies constraints on epistemic-state 
inference without selecting a particular implementing algorithm. 
Second, the accounts infer different objects. Zhang and colleagues 
infer reward-function parameters (preferences and movement costs) 
within an experimenter-specified gridworld MDP. The agent's epistemic 
access to the relevant task state --- its beliefs and percepts --- is 
assumed. ToM-U instead derives belief-like states from ordered 
information-access histories and source credibility. It is therefore 
situated upstream of utility-inference accounts, supplying 
epistemic-state estimates that could condition their inference of 
preferences and costs. Although Zhang and colleagues integrate multiple 
observations in Experiment~3, their sequential condition carries the 
posterior from one observation forward as the prior for the next. It 
does not represent the source-, trust-, and order-sensitive epistemic 
provenance encoded by \(H_i\).

Third, the accounts make contrasting commitments about the 
representation of uncertainty. Zhang and colleagues propose that 
mentalizing maintains a posterior distribution over latent 
mental-state parameters, with uncertainty represented by distributional 
spread. ToM-U instead proposes that mentalizing terminates in a 
discrete, categorical LEWM selected through generate-and-filter 
inference, with residual uncertainty represented by scalar confidence. 
This constitutes an empirically distinguishable contrast 
(\S~\ref{sec:predictions}). Fourth, Zhang and colleagues identify 
recursive mentalizing (reasoning about another agent's reasoning about 
oneself) as an open problem. The proliferation mechanism in Definition~3 
directly formalizes that case. The two accounts are therefore 
complementary: a utility-inference model of the kind Zhang and 
colleagues propose could operate downstream of, and condition its 
inference on, the belief-state estimates produced by ToM-U.

A more limited structural parallel appears in cognitive hierarchy theory
(CHT): both frameworks use discrete depth levels to model bounded
reasoning \citep{camerer2004cognitive}. Their inferential targets differ,
however. ToM-U concerns epistemic-state inference, whereas CHT concerns
action choice under strategic uncertainty.

Active-inference accounts provide another point of contact
\citep{friston2017active}. They specify a variational mechanism operating
under a generative model; ToM-U instead constrains the
epistemic-inference problem that such a mechanism could implement. The
accounts may therefore be complementary.

The generate-and-filter architecture of Definition~4 shares structural 
similarity with abductive inference: generating candidate explanations 
and evaluating them for coherence with available evidence rather than 
computing likelihoods over a distribution 
\citep{hobbs1993interpretation,thagard1989explanatory}. Fit scoring 
with its scalar confidence accumulator could be interpreted as a 
type of coherence evaluation specific to mentalizing. 

ToM-U occupies an unusual position in the broader debate around how 
people mentalize, largely thanks to its formal epistemic state inference 
apparatus. The two main representational accounts, simulation 
theory~\citep{gordon1986folk,goldman2006simulating} and 
theory-theory~\citep{wellman1990child,gopnik92}, are not formalized. 
Coincidentally, formalizing the information access history problem 
produced a theory that reflects both traditions. For example, the 
generative logic of ToM-U's third-person hypothesis construction about 
another agent's epistemic state, versus self-as-model, nods to 
theory-theory. Meanwhile, its evaluate-against-behavior structure in 
which candidates are evaluated by how well they account for observations 
acknowledges simulationism. 

Its clearest departure from both traditions is the treatment of nested
selves as constructed others in situations demanding recursive reasoning
of the \textit{I-know-that-you-know-that-I-know...} type. The projected self 
at each proliferation depth of such reasoning is a constructed other 
reasoned about from the outside. The second ``I'' is not the focal agent 
running its own machinery in a transformed mode. This treatment follows 
directly from the proliferation architecture.

A further distinction is the LEWM itself: a temporary,
occasion-specific working model of another agent's epistemic situation,
constructed from information-access history, evaluated for coherence
with observed behavior, and adopted or discarded as the operative
account of that situation. ToM-U posits that mentalizing is accomplished 
by generating and evaluating these working models, not through a process 
analogous to scientific hypothesis testing or by simulating another mind. 

The discrete-LEWM commitment introduces a related distinction:
sufficiently understanding a target's BLSs to act naturally yields a
categorical outcome. Although confidence is graded, the focal agent's 
LEWM is a commitment, not a distribution. This maps onto Dennett's 
intentional stance (\citeyear{dennett1987intentional}): attributing 
beliefs to an agent is inherently a categorical act, one either adopts 
the intentional stance or not, even if the underlying processes 
supporting that attribution are graded or probabilistic 
\citep{sanborn2016bayesian}. 

The termination rule interacts with the residue-attribution rule. Under
the implementation examined here, a stable weakest direction determines 
rejection under the \(\min\) rule and is treated as the implicated 
relationship; repeated invocations therefore concentrate residue on the 
same edges. The weighted-average rule instead implicates every 
pair short of full coherence, spreading residue across their dependency 
edges. Because the persistence floor preserves part of accumulated residue, 
this divergence can remain after acute residue decays. The simulation 
demonstrates these conditional consequences for one asymmetric 
multi-agent configuration (Appendix \ref{supp:run2}).

\subsection{Empirical Predictions of ToM-U}
\label{sec:predictions}

An agent’s information-access history constrains the candidate LEWMs 
available to the focal agent. Overestimating a target’s exposure can 
admit belief attributions the represented history does not support; 
underestimating exposure can exclude otherwise supported attributions. 
Source-monitoring errors should therefore bias belief attribution in 
the direction of the exposure error \citep{johnson1993source}. For 
example, if the observer did not know that the testimony of the popcorn 
bag's contents came from a trusted source, then they would produce a 
different LEWM than an observer who knows Sam trusts the source. More 
broadly, the focal agent's past relationship history with the target, 
including trust history and accumulated mentalizing residue, is encoded 
in the edge structure as trust-typed BLSs and accumulated cred 
reductions, which shapes candidate generation. An observer with no past 
relationship to Sam has a different and likely impoverished candidate 
space than one with an established trust history.

LEWM construction errors are distinct from observability-driven errors. 
The observability function, obs, dictates the clarity of the 
relationships in the LEWM at invocation. Poor observability can lower 
confidence or trigger rejection even when the selected candidate’s 
content is correct. Because observability also constrains reconstructed 
information histories, it can produce content errors as well. ToM-U 
therefore distinguishes confidence effects caused by weak access from 
candidate-space effects caused by missing or distorted exposure 
information.

In ToM-U, BLS-type attribution is constrained by the graph structure,
\(\text{obs}\), \(H_i\), and agent characteristics rather than treated
as an unconstrained labeling task. A LEWM in which Sam \textit{wants}
the bag to contain popcorn is therefore distinct from one in which she
\textit{believes} it does.

The typed-edge representation also supports several hypotheses
conditional on how candidate generation is implemented. Under an
accessibility-biased generator, weak evidence for the correct BLS type
should increase selection of contextually accessible types. For
example, an observer lacking evidence that Sam wants popcorn might
default to a belief-labeled edge. An implementation permitting
cross-type coupling could also allow an overconfident attribution in
one BLS type to contaminate estimates in another, as in a
curse-of-knowledge effect \citep{camerer1989curse}. Finally, if lower
sophistication limits the ability to maintain distinct typed candidates,
mentalizing sophistication should correlate with the ability to
dissociate belief and desire attribution, consistent with the
developmental literature \citep{wellman1990simple}. The current
definitions make these effects representable but do not independently
entail the corresponding implementation rules.

Another failure mode arises when the focal agent miscalibrates a
target's sophistication. Because the branch rooted at \(a_i\) is bounded
by \(S_i^{\mathrm{eff}}=\min(S_1,S_i)\), underestimating \(S_i\) lowers
the maximum depth attributed to the target and can prematurely truncate
recursive reasoning that the focal agent is otherwise capable of
representing. Overestimating \(S_i\) raises the target-relative ceiling
and can permit a deeper attributed structure than the target actually
uses, although proliferation may still halt earlier because of
branching-width or marginal-gain constraints. Miscalibration should
therefore bias the complexity of the constructed tree in different
directions: underestimation can omit nested states relevant to the
target's behavior, whereas overestimation can introduce unnecessarily
elaborate nested states. This prediction aligns with evidence that
people systematically underestimate the depth of others' strategic
reasoning \citep{camerer2004cognitive}.

The graph structure also distinguishes failures involving absent and
false beliefs. A child who fails the Wimmer and Perner task may have no
representation of the target's belief state at all or may represent that
state but attribute the wrong one
\citep{wimmer83,pillow1989early,pratt1990young,wellman2001meta}.
These possibilities correspond to structurally different failures:
absent belief involves no relevant edge in \(H_i\), whereas false belief
involves an existing belief-typed edge to a content node that does not 
match the world.

More broadly, as a formalized theory ToM-U is falsifiable via its 
core architectural commitments. Concretely:
\begin{itemize}
    \item Systematic evidence that manipulating the order or provenance 
    of information leaves belief attribution unchanged, under conditions 
    in which ToM-U predicts different candidate LEWMs, would count 
    against the LEWM’s representational foundation.
    \item Evidence that mentalizing obligatorily maintains 
    distributions over candidate mental states rather than selecting an 
    operative candidate with scalar confidence would conflict with the 
    generate-and-filter architecture.
    \item Evidence that rejected mentalizing attempts produce no 
    relationship-specific carryover under controlled repeated-inference 
    conditions would count against the residue function. 
\end{itemize}
Each of these architectural commitments is amenable to empirical testing.

\subsection{Motivated Reasoning and ToM-U}
ToM-U identifies three loci at which motivated reasoning could influence 
mentalizing: information-access history, represented state information, 
and estimated observability. The current formalization treats these 
quantities as inputs rather than modeling their distortion, so the three 
pathways support implementation-dependent hypotheses rather than direct 
architectural predictions. First, motivated reasoning impacts what 
information we attend to, retain, and the merit we give it 
\citep{kunda90}. In these ways, it can shape information-access histories 
on both sides of the inference. Distortions in a target's history carry 
into the focal agent's reconstruction of \(H_i\), while distortions in 
the focal agent's own \(H_1\) can produce motivated commitments that 
constrain candidate generation, as in the motivated variant of the worked 
example. Second, the focal agent is also subject to motivated reasoning, 
like wishful thinking, that can distort what they observe, meaning \(O\) 
might carry biased affective or motivational information that corrupts 
backward inference. Third, during self-projection, motivated reasoning 
can distort the focal agent's estimates of how observable their own 
mental states are to the target, systematically biasing the obs scalars 
on relevant edges. Such miscalibration of obs estimates can potentially 
serve as a substrate for self-deception as seen in the illusion of 
transparency \citep{gilovich1998illusion} or spotlight effect 
\citep{gilovich2000spotlight}. More broadly, systematic miscalibration 
of obs estimates may contribute to mentalization failures documented 
in clinical populations \citep{fonagy2002affect}.

\subsection{Domain Generality and Extensions}

The formal machinery of ToM-U fits with communications theory. The 
generate-and-filter procedure aligns with relevance theoretic accounts 
of utterance comprehension, and ordered \(H_i\) functions analogously to 
the cognitive environment construct, or the information an agent has 
access to at a given moment \citep{sperber1986relevance}. At the level 
of joint understanding, mutual reconciliation in Definition~4 formalizes 
the grounding process described by Clark and Brennan 
(\citeyear{clark91}), in which mutual understanding is achieved through 
iterated exchange, not simultaneous joint inference.

The \(H_i\) framework is well positioned for modeling memory for social 
interactions, particularly source memory: tracking who said what to 
whom, and in what order, is essentially \(H_i\) with temporal indexing 
\citep{johnson1993source}. Similarly, snapshot timestamps and 
proliferation tree pruning connect to event segmentation theory in which 
event boundaries update situation models \citep{zacks2007event}. 

ToM-U produces epistemic-state estimates for downstream consumption. A
higher-order parent process can use those estimates to support social
outcomes such as coordination, cooperation, or conflict avoidance. 
Because the utility is domain-agnostic, it can also be invoked by 
processes not directed toward an immediate instrumental outcome; for 
example, curiosity pointed at social observation might give rise to 
spontaneous ToM \citep{apperly2009humans,gurney2024spontaneous}. It is 
also worth noting speculatively that cognitive hierarchy models in game 
theory, which presuppose that agents reason about other agents' 
strategic reasoning to some bounded depth, may benefit from 
ToM-U style epistemic state estimates as inputs, another opportunity 
for future theory-driven work \citep{camerer2004cognitive}.

\subsection{Limitations and Extensions}
Each of ToM-U's parameters warrants contextual and cultural calibration. 
Cross-cultural work documents variation in the timing of false belief 
task success across cultures \citep{callaghan2005synchrony} as well as 
differences in the sequence of ToM development 
\citep{shahaeian2011culture}, suggesting that parameters such as 
\(S_{\max}\) and \(\tau_{\text{high}}\) may require cultural adjustment. 
Likewise, the cognitive hierarchy literature documents substantial 
variation in sophistication levels, \(S_i\), across contexts and 
populations \citep{camerer2004cognitive}.

Individual differences in mentalizing ability also offer a promising 
avenue for future empirical research. Key sources of variation --- 
such as sophistication calibration, mentalizing style, and sensitivity 
to residue accumulation --- could be especially informative when 
studied in populations with documented mentalization difficulties 
\citep{fonagy2002affect}. More broadly, ToM-U was developed as a theory 
of adaptive mentalizing. Maladaptive cases that give rise to 
self-deception and related clinical phenomena may gain insight from 
ToM-U, with miscalibration of the observability scalar representing a 
natural formal substrate for modeling clinical mentalization failures. 

ToM-U can be extended along several formal dimensions to expand its 
descriptive scope. First, giving \(H_i\) veridicality weights would enable 
modeling confabulation and reconstructive memory 
\citep{loftus1974reconstruction,schacter1999seven}. Second, a means of 
abandoning a LEWM mid-construction while still accumulating residue 
would extend the model to capture a broader class of inference 
failures. Third, further developing the multi-agent architecture of 
mutual reconciliation is warranted, as it provides a natural framework 
for modeling collective epistemic dynamics and distributed sensemaking. 
Lastly, the residue function's accumulation dynamics may also provide a 
formal substrate for modeling hypervigilance in interpersonal contexts, 
with repeated rejection events on specific edge types producing 
heightened mentalizing sensitivity.

Beyond these extensions, ToM-U leaves the functional forms of fit scoring, 
confidence accumulation, candidate generation, \(F(e)\), the \(R\) to cred 
mapping, and residue decay open. Specifying and empirically comparing 
candidate functions for these open components is a natural next step.

\section{Conclusion}
The Theory of Mind Utility formalizes how a person infers epistemic
states when reasoning about other minds. This distinct cognitive problem
is left representationally underspecified by accounts such as
theory-theory and simulation theory and generally treated as an input by
formal accounts such as Bayesian ToM and the partially observable Markov
decision process lineage in computer science. The five definitions keep 
ToM-U’s representational and inferential machinery independent of any 
particular task domain and support directional predictions about failure. 
Understanding the full range of social cognitive phenomena, from false 
belief to motivated opacity to collective sensemaking, necessitates a 
formal account of how minds construct candidate models of other minds; 
the Theory of Mind Utility addresses this need. 

\section*{Acknowledgments}

The views and conclusions contained in this document are those of the 
authors and should not be interpreted as representing the official 
policies, either expressed or implied, of the Army Research Office or 
the U.S. Government. The U.S. Government is authorized to reproduce and 
distribute reprints for Government purposes notwithstanding any 
copyright notation herein.

\section*{Declarations}

\subsection*{Funding}

This work was supported by the Army Research Office under Cooperative
Agreement Number W911NF-25-2-0040.

\subsection*{Competing interests}

The authors have no competing interests.

\subsection*{Data, materials and code availability}

No empirical participant data were collected for this study. The 
reference implementation, scripts, and machine-readable outputs 
supporting the simulation demonstrations are available from the 
first author.

\subsection*{Ethics approval}

Not applicable. This article does not report research involving human
participants or animals.

\subsection*{Consent to participate}

Not applicable.

\subsection*{Consent to publish}

Not applicable.

\subsection*{Author contributions}

Nikolos Gurney conceptualized the theory, acquired funding, and drafted 
and edited the manuscript. Stacy Marsella contributed to the 
conceptualization and edited the manuscript. Both authors read and 
approved the final manuscript.

\begin{appendices}

\section{Reference Simulation as a Consistency Witness}
\label{supp:sim}

\subsection{Scope and status}
\label{supp:scope}
The reference simulation is a \emph{consistency witness} rather than an
empirically fitted model. It establishes that the five definitions
compose into a runnable system and that the boundary conditions of
Definition~5 are jointly satisfiable. To do so, the implementation fixes
illustrative forms for candidate generation, fit scoring, confidence 
accumulation, observability filtering, tree rebuilding and pruning, 
projected-node coherence, residue dynamics, and the motivated-commitment 
rule of \S\ref{supp:motivated}. These choices are not theoretical 
commitments. Accordingly, the witness tests compositional coherence, not 
parameter robustness or empirical fit. Its role is to pay ToM-U's coherence 
and instantiability debt; parameters are fixed within each analysis 
(Table~\ref{supp:tab:params}).

\begin{table}[htbp]
\centering
\caption{Default parameters of the reference witness and reported runs.}
\label{supp:tab:params}
\begin{tabular}{@{}llll@{}}
\toprule
Parameter & Value & Parameter & Value \\
\midrule
$\tau_{\text{high}}$ & $0.90$ & $\lambda$ (residue decay) & $0.10$ \\
$\tau_{\text{low}}$ & $0.50$ & $\alpha$ (acute increment) & $0.50$ \\
$\tau_{\text{commit}}$ & $0.70$ & $\beta$ (floor increment) & $0.20$ \\
$\rho$ (branch decay) & $0.50$ & $\gamma$ (rehabilitation) & $0.30$ \\
$\kappa$ (proliferation cost) & $0.05$ & discount factor & $0.10$ \\
$S_{\max}$ / $S_1$ / target $S_i$ & $6$ / $3$ / $3$ & inter-session interval $\Delta t$ & $1.0$ \\
belief-source strength & $0.50$ & untyped-source strength & $0.50$ \\
\bottomrule
\end{tabular}
\end{table}

The isolated depth-two compositionality check fixes \(S_1=S_2=3\), so 
\(S_2^{\mathrm{eff}}=3\), and fixes \(\rho=0.8\), \(b(2,3)=1\), and
\(\tau_{\text{high}}=0.99\), so that the effect of the projected node 
remains visible in the terminal classification. The discount factor 
scales the evidential weight of a source that another source instructs 
the target to disregard (the testimony's ``ignore the label'' relation; 
Run~1). Belief-source and untyped-source strengths are the default 
evidential weights of sources without a trust-typed edge. Scenario 
inputs: the testimony is trust-typed with trust \(0.90\) and 
observability \(0.85\); the label has observability \(1.0\). In Run~2, 
the focal agent's observability of \(a_2\), \(a_3\), and \(a_4\) is 
\(0.55\), \(0.20\), and \(0.40\). All other values retain their 
defaults. 

\subsection{Implementation}
\label{supp:impl}

The implementation contains one module corresponding to each definition: 
Definition~1 (LEWM), Definition~2 (agent node; ordered \(H_i\)),
Definition~3 (bounded proliferation), Definition~4 (backward inference, 
self-projection, mutual reconciliation), Definition~5 (residue). Belief 
content is represented as a belief-typed edge to a content state node 
(e.g.\ \(a_2 \to o_{\text{pickle}}\)), never as a value on an 
agent-to-agent edge. A focal agent's candidate selection and inference 
confidence are held in a separate record, off the graph. The current 
record identifies the operative selection, while an ordered record 
history preserves earlier write-backs; residue-bearing relationships are 
retained. Residue targets are derived from each candidate's dependency 
edges. 
Every reported scenario places at most one typed edge on each ordered 
pair, so the implementation keys edges by their endpoints; this is a 
special case of Definition~1, not a departure from it.
Because the witness enumerates a finite set of candidates, 
inference ends after all available configurations have been evaluated 
even when this occurs before the sophistication ceiling; the ordinary 
confidence thresholds then classify the result. The model, executable 
checks, and TXT/CSV generation use only the Python standard library; 
Matplotlib is optional and generates the vector-PDF figure. The 
implementation is deterministic, and repeated runs of a given output are 
identical. 

\subsection{Consistency of Definition~5 (the witness)}
\label{supp:witness}

\noindent\textbf{Proposition (Consistency of Definition~5).}
\emph{For any credibility mapping \(\mu:[0,\infty)\to(0,1]\) that is 
strictly decreasing with \(\mu(0)=1\), there exists 
a residue function \(R\) satisfying boundary conditions~1--3, 5, and~6 
of Definition~5; conditions~4 and the credibility clause of condition~5 
then hold for every such \(\mu\).} The witness exhibits one such pair, 
\(R(e,t)=D(e)\,e^{-\lambda(t-t_{\text{last}})}+F(e)\)
and \(\mu(r)=1/(1+r)\).

At a rejection event of degree \(d\) after elapsed time \(\Delta t\),
the witness updates:
\begin{equation}
D(e) \leftarrow D(e)e^{-\lambda\Delta t}+\alpha d,
\qquad
F(e) \leftarrow \min\{1,F(e)+\beta d\}.
\end{equation}
At a supporting event of strength \(\sigma\), it updates:
\begin{equation}
D(e) \leftarrow \max\{0,D(e)e^{-\lambda\Delta t}-\gamma\sigma\},
\end{equation}
while leaving \(F(e)\) unchanged. Each event resets \(t_{\text{last}}\)
to the event time.

An executable check confirms all six conditions on the witness
(Table~\ref{supp:tab:witness}).

\begin{table}[htbp]
\centering
\caption{Executable verification of Definition~5's six boundary conditions.}
\label{supp:tab:witness}
\begin{tabular}{@{}p{0.43\linewidth}p{0.42\linewidth}c@{}}
\toprule
Condition & Check & Result \\
\midrule
Targeting & untouched edge: $R=0$, $\text{cred}=1$ & PASS \\
Monotonicity in disconfirmation & larger $d \to$ larger $R$ & PASS \\
Monotonicity in time & $R$ non-increasing between events & PASS \\
Asymptotic lower bound on credibility & $\text{cred}>0$ after finite rejections & PASS \\
Persistence & $R \ge F > 0$; full rehabilitation lands at $F$ & PASS \\
Partial rehabilitability & support lowers $R$, not below $F$ & PASS \\
\bottomrule
\end{tabular}
\end{table}

The floor-plus-decay decomposition is what makes conditions~3 and~5 
jointly satisfiable. Elapsed time drives \(R\) toward \(F\), rather than 
toward zero, so acute residue can diminish without eliminating the 
persistent effect of a rejected model.

\subsection{Demonstrations}
\label{supp:runs}

\subsubsection{Source-typing and source-discount (Run~1)}
\label{supp:run1}

Backward inference on the trusted-testimony variant, crossing whether 
source discounts are applied (the testimony's explicit ``ignore the 
label'' relation) with whether sources are trust-typed 
(Table~\ref{supp:tab:run1}). The ToM-U claim is the diagonal of the 
\(2 \times 2\): the full model recovers popcorn, while the model 
stripped of both commitments reproduces the surface-signal failure, 
fixating on the most observable source (chocolate). Removing either 
commitment alone still recovers popcorn, as each is independently 
sufficient here. This instantiates the contrast with surface-signal 
systems established in \S\ref{sec:popcorn} and illustrated in 
\S\ref{worked:nodes}. Fit in the witness is insensitive to order, 
so Run~1 tests source typing and discounting, not the order 
sensitivity of \(H_i\).

\begin{table}[htbp]
\centering
\caption{Run~1. Attribution and terminal confidence by ablation.
$*$ marks the headline diagonal.}
\label{supp:tab:run1}
\begin{tabular}{@{}cclll@{}}
\toprule
discounts & typed & attribution & confidence & classification \\
\midrule
T & T & popcorn   & $0.942$ & ACCEPTED\quad$*$ \\
T & F & popcorn   & $0.906$ & ACCEPTED \\
F & T & popcorn   & $0.761$ & ACCEPTED\_LOW \\
F & F & chocolate & $0.752$ & ACCEPTED\_LOW\quad$*$ \\
\bottomrule
\end{tabular}
\end{table}

\subsubsection{Reconciliation and residue divergence (Run~2)}
\label{supp:run2}

To make the consequences of the two joint-confidence rules visible, the 
simulation imposes one controlled asymmetry: the focal agent has less 
access to the history of one agent (\(a_3\)) than to the histories of 
two other targets. This lower observability is part of the scenario 
design, not something inferred from \(a_3\)'s behavior; it ensures that 
the \(a_3\) pair remains difficult to reconcile while the other two 
pairs are comparatively well supported.

\begin{figure}[htbp]
\centering
\includegraphics[width=\linewidth]{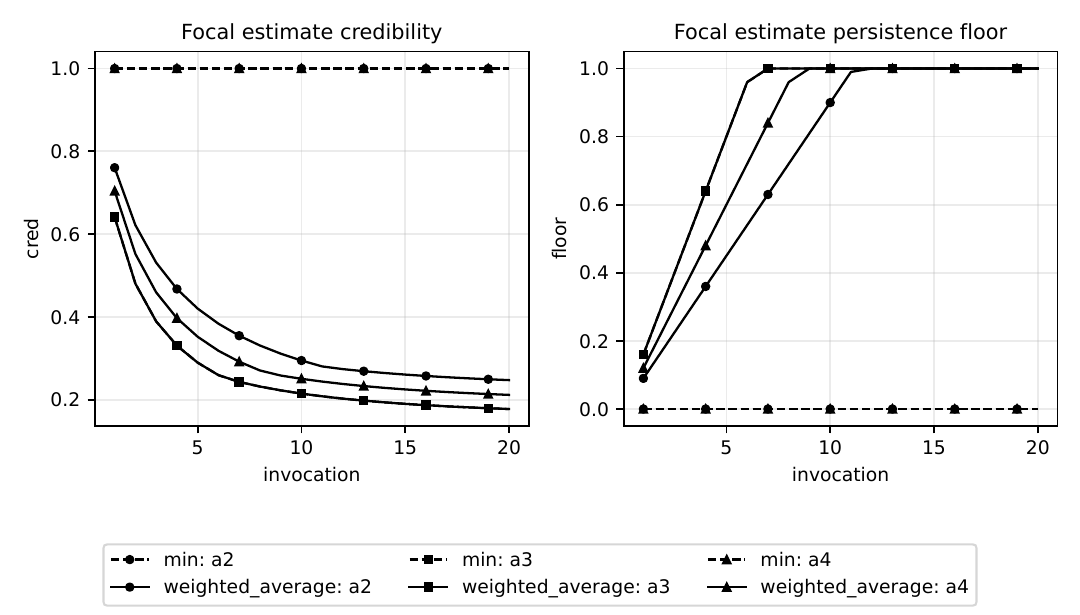}
\caption{Per-edge credibility and persistent residue floor across twenty
invocations under minimum and weighted-average reconciliation.}
\label{supp:fig:run2}
\end{figure}

Mutual reconciliation is then run for twenty invocations under each 
joint-confidence rule. Both rules reject every invocation, so both 
accrue residue; they differ in \emph{where} it accrues 
(Table~\ref{supp:tab:run2}; Figure~\ref{supp:fig:run2}). 
Under \(\min\), only the single weakest 
pair is implicated each round, so residue concentrates on the \(a_3\) 
edges; under the weighted average, every pair short of full coherence is 
implicated, so it spreads across all six reported dependency edges. 
Before any floor saturates, the weighted-average rule still assigns the 
\(a_3\) pair a disproportionate share, because disconfirmation is 
largest for the weakest pair. The minimum formulation is the rule 
specified by \S\ref{def:inference}. The alternative weighted-average 
formulation (equal weights) embodies the different commitment described 
in \S\ref{dis:theoryPosition}: coherence is distributed across agent pairs 
and partial alignment is aggregable. Rejection residue is applied at the 
terminal reconciliation-round timestamp. This makes concrete the 
``qualitatively different residue profile'' of \S\ref{dis:theoryPosition}.

\begin{table}[htbp]
\centering
\caption{Run~2. Residue concentration versus spread across invocations. 
``Floored edges'' counts edges with \(F(e)>0\) (constant from the first 
invocation); ``weakest-pair share'' is the \(a_3\) pair's fraction of 
total floor. Floors are capped at 1 in the witness; all implicated edges 
reach the cap by invocation 12, after which the weighted-average share 
equals \(1/3\) by construction.}
\label{supp:tab:run2}
\begin{tabular}{@{}lcccc@{}}
\toprule
& & \multicolumn{3}{c}{Weakest-pair share at invocation} \\
\cmidrule(l){3-5}
Termination rule & Floored edges & 5 & 10 & 20 \\
\midrule
$\min$ & $2$ & $1.00$ & $1.00$ & $1.00$ \\
Weighted average (equal weights) & $6$ & $0.43$ & $0.35$ & $0.33$ \\
\bottomrule
\end{tabular}
\end{table}

\subsubsection{Faithfulness across the four variants (Run~3)}
\label{supp:run3}

Run~3 verifies that the implementation reproduces the worked example. It 
confirms the model distinguishes \emph{absent} belief from \emph{false} 
belief, and it exercises the residue-caused motivated episode described 
next.

\begin{table}[htbp]
\centering
\caption{Run~3. The four variants through the full pipeline. For the 
motivated variant the row reports the terminal episode outcome; 
``floored'' indicates residue on the \(a_2 \to o_{\text{pickle}}\) edge.}
\label{supp:tab:run3}
\begin{tabular}{@{}lllll@{}}
\toprule
Variant & Attribution & Confidence & Classification & Floored \\
\midrule
standard  & chocolate & $1.000$ & ACCEPTED & no \\
trusted   & popcorn   & $0.942$ & ACCEPTED & no \\
naive     & absent    & $1.000$ & ACCEPTED & no \\
motivated & popcorn   & $0.942$ & ACCEPTED & yes \\
\bottomrule
\end{tabular}
\end{table}

\subsubsection{The motivated episode: residue-caused fallback}
\label{supp:motivated}

The motivated variant models an observer with a preference for a reality 
in which Sam believes the bag contains a pickle, encoded as a 
belief-typed edge \(a_2 \to o_{\text{pickle}}\). A session is 
\emph{committed} to that hypothesis while the edge's (decay-adjusted) 
credibility is at least \(\tau_{\text{commit}}\). A committed reasoning 
session considers only pickle-consistent LEWMs. Here, three 
configurations differing in the attributed credence (BLS scalars \(0.6, 
0.8, 1.0\)) are implemented. Each fails to cohere with Sam's information 
access history, which contains popcorn and chocolate evidence but 
provides no support for pickle and favors popcorn through the trusted 
testimony. The session reaches the sophistication ceiling with 
confidence below \(\tau_{\text{low}}\) and is rejected, accruing residue 
on \(a_2 \to o_{\text{pickle}}\) and lowering its credibility below 
\(\tau_{\text{commit}}\). The next session is therefore uncommitted: it 
considers the pickle hypothesis alongside the others but no longer 
exclusively, and the evidence-coherent popcorn account wins, returning 
popcorn with high confidence (Table~\ref{supp:tab:episode}).

\begin{table}[htbp]
\centering
\caption{The motivated episode. Residue on \(a_2 \to o_{\text{pickle}}\) 
ends the committed pursuit; the fallback session adopts the 
evidence-coherent belief.}
\label{supp:tab:episode}
\begin{tabular}{@{}lllll@{}}
\toprule
Session & Committed & Attribution & Confidence & Outcome \\
\midrule
1 & yes & pickle   & $0.000$ & REJECTED ($\ell = S_1 = 3$) \\
2 & no  & popcorn  & $0.942$ & ACCEPTED \\
\bottomrule
\end{tabular}
\end{table}

Residue \emph{causes} the fallback. It ends the motivated exclusivity, 
not by lowering popcorn's fit but by preventing renewed commitment to 
pickle at the next session. Two checks establish this. A counterfactual 
in which the session-1 residue update is suppressed leaves credibility 
above \(\tau_{\text{commit}}\), so both tested sessions remain committed 
and rejected and the fallback never occurs. Under the default parameters 
this is a next-session fallback, not necessarily permanent abandonment. 
In the absence of further residue or evidence, the acute component 
decays and credibility approaches approximately \(0.833\), above the 
default commitment threshold, so later recommitment remains possible.

A limiting case sets \(\tau_{\text{commit}} = 0\): because Definition~5 
(\S\ref{def:residue}) keeps credibility strictly positive after any finite 
history, commitment is retained across a fixed ten-session horizon and 
the hypothesis is not abandoned within that horizon. The adaptive 
default and this non-abandoning limit are separated by a single 
parameter, the formal counterpart of the difference between ordinary 
belief revision and perseverative attribution. 

\subsubsection{Compositionality}
\label{supp:composition}

A structural check confirms that Definition~3's 
(\S\ref{def:proliferation}) proliferation is consumed by Definition 4 
(\S\ref{def:inference}). At depth two, the observer represents Sam as 
representing her friend as believing popcorn, coherently matching the 
friend's testimony. This configuration returns joint confidence \(0.942\)
and is ACCEPTED\_LOW under the check's fixed 
\(\tau_{\text{high}}=0.99\). 
Perturbing the projected friend's belief to chocolate makes it conflict 
with the testimony, reducing joint confidence to \(0\) and producing 
REJECTED. A separate check confirms that self-projection independently 
selects from a target-filtered projection of \(H_1\), instead of 
inheriting the content selected by backward inference. Taken together 
with the residue witness and reported runs, these checks establish that 
the five definitions compose operationally.

\subsection{Traceability}
\label{supp:trace}
Table~\ref{supp:tab:trace} maps each artifact reported above to the 
claim it instantiates. No artifact is offered as empirical support; 
each shows that a stated commitment is realizable in a running system.
\begin{table}[htbp]
\centering
\caption{Each artifact and the claim it instantiates.}
\label{supp:tab:trace}
\begin{tabular}{@{}p{0.34\linewidth}p{0.62\linewidth}@{}}
\toprule
Artifact & Instantiates \\
\midrule
Witness (\S\ref{supp:witness}) & Definition~5 six-condition satisfiability \\
Run~1 (\S\ref{supp:run1}) & source-typing/discount vs.\ surface-signal failure \\
Run~2 (\S\ref{supp:run2}) & $\min$ vs.\ weighted average $\to$ concentrated vs.\ spread residue \\
Run~3 (\S\ref{supp:run3}) & faithfulness; absent vs.\ false belief \\
Motivated episode (\S\ref{supp:motivated}) & residue-caused fallback; adaptive vs.\ perseverative regimes \\
Compositionality (\S\ref{supp:composition}) & all five definitions compose operationally \\
\bottomrule
\end{tabular}
\end{table}
\end{appendices}

\clearpage
\bibliographystyle{apalike}
\bibliography{bibliography}

\end{document}